\pdfoutput=1

\documentclass[11pt]{article}
\usepackage{adjustbox}
\usepackage[final]{acl}
\usepackage{times}
\usepackage{latexsym}
\usepackage{amsmath}
\usepackage{booktabs}
\usepackage{multirow}
\usepackage{float}
\usepackage{listings}
\usepackage{adjustbox}
\usepackage{amsmath}
\usepackage{amssymb}  

\usepackage[T1]{fontenc}

\usepackage[utf8]{inputenc}

\usepackage{microtype}

\usepackage{inconsolata}

\usepackage{graphicx}

\title{Evaluating NL2SQL via SQL2NL}

\author{
Mohammadtaher Safarzadeh \quad Afshin Oroojlooyjadid \quad Dan Roth \\
Oracle AI \\
\texttt{\{mohammadtaher.safarzadeh, afshin.oroojlooyjadid, dan.roth\}}@oracle.com
}
  
\begin{document}

\maketitle

\begin{abstract}
Robust evaluation in the presence of linguistic variation is key to understanding the generalization capabilities of Natural Language to SQL (NL2SQL) models, yet existing benchmarks rarely address this factor in a systematic or controlled manner. We propose a \textit{novel schema-aligned paraphrasing framework} that leverages SQL-to-NL (SQL2NL) to automatically generate semantically equivalent, lexically diverse queries while maintaining alignment with the original schema and intent. This enables \textit{the first targeted evaluation} of NL2SQL robustness to linguistic variation in isolation---\textit{distinct from prior work} that primarily investigates ambiguity or schema perturbations. Our analysis reveals that state-of-the-art models are far more brittle than standard benchmarks suggest. For example, LLaMa3.3-70B exhibits a 10.23\% drop in execution accuracy (from 77.11\% to 66.9\%) on paraphrased Spider queries, while LLaMa3.1-8B suffers an even larger drop of nearly 20\% (from 62.9\% to 42.5\%). Smaller models (e.g., GPT-4o mini) are disproportionately affected. We also find that robustness degradation varies significantly with query complexity, dataset, and domain---\textit{highlighting the need for evaluation frameworks that explicitly measure linguistic generalization} to ensure reliable performance in real-world settings.
\end{abstract}

\section{Introduction}
\label{sec:intro}

A large share of enterprise data reside in relational databases, with SQL as the primary query interface. Machine learning models that translate natural language (NL) to SQL (NL2SQL) have become essential for seamless human-database interaction. However, evaluating these models remains a significant challenge. Public benchmarks such as Spider~\citep{Spider_paper} and BIRD~\citep{Bird_paper} often oversimplify real-world complexities, overlooking factors such as linguistic diversity, schema variations, and domain-specific constraints. As a result, reported model performance on these datasets may not reflect true generalization capabilities.

To address these limitations, the field has called for fine-grained evaluation methods that go beyond broad benchmarks. Such methods should assess model performance across varying query complexities, identify specific failure points, and support targeted improvements. Without detailed evaluations, critical weaknesses may remain hidden, limiting the robustness and reliability of NL2SQL models in practical applications.

Robustness remains a persistent challenge for NL2SQL systems. Robustness encompasses the ability to handle linguistic and structural perturbations, including ambiguous, paraphrased, or complex queries. For instance, \citet{bhaskar-etal-2023-benchmarking} introduced AmbiQT, a dataset targeting lexical and structural ambiguity, and proposed LogicalBeam, a decoding algorithm that increases SQL diversity and improves match accuracy. Similarly, \citet{wang-etal-2023-know} systematically categorized ambiguous and unanswerable queries, using counterfactual generation to strengthen model performance. Broader robustness evaluations, such as those by \citet{dr_spider}, have revealed that state-of-the-art models can suffer significant accuracy drops under both database and NL question perturbations, highlighting their fragility in real-world scenarios.

A central source of this fragility is schema linking, the process of identifying and aligning schema elements (column and table names) with their references in NL queries. Schema linking is widely recognized as pivotal to NL2SQL performance~\citep{guo2019, bogin2019, wang2020, chen2020, cao2021}, with improvements directly enhancing parsing accuracy~\citep{lei2020}. However, schema linking remains brittle, particularly when faced with synonym substitutions or paraphrased natural language queries~\citep{gan2021a}. Recognizing these challenges, Spider 2.0~\citep{lei2025spider} was recently introduced to better reflect real-world use cases and address the limitations of earlier benchmarks.

In this work, we present a comprehensive evaluation framework for NL2SQL models that captures the nuanced challenges of robustness, with a particular emphasis on schema linking. Our approach enables fine-grained analysis across query complexities and ambiguity types, providing actionable insights for diagnosing model failures and assessing generalization. Unlike prior work that focuses on improving NL2SQL model performance through architectural innovations or pre-training strategies, our contribution lies in systematically characterizing model behavior. For example, \citet{xu-etal-2018-sql} proposed a graph-to-sequence model to better capture SQL’s structural patterns, while GraPPa~\citep{yu2021grappa} enhanced compositional generalization using schema-aware pre-training on synthetic data. In contrast, our framework is designed to surface detailed evaluation signals that can inform the development or fine-tuning of NL2SQL models.

\subsection{Our Approach}
Understanding the root causes of errors in predicted SQL queries is a challenging and entangled task. Model failures can arise from multiple sources—such as incorrect schema linking, misinterpretation of linguistic cues, or logical inconsistencies—making it difficult to diagnose performance breakdowns or meaningfully compare model behaviors. A prerequisite to effective error analysis is the ability to isolate specific sources of error.
To this end, we introduce an automatic natural language query generation framework that embeds schema alignment directly into the generated queries. \textbf{By construction, SQL2NL decouples the impact of schema linking from other factors, enabling a focused evaluation on linguistic variation} and model generalization. This design innovation serves two primary objectives:
\begin{enumerate} \item \textbf{Isolating performance degradation due to linguistic variation} by holding schema alignment constant—facilitating more precise analysis of model robustness. \item \textbf{Generating high-quality, schema-consistent preference datasets} for training and fine-tuning NL2SQL models in a controlled and interpretable manner. \end{enumerate}
Unlike prior methods that rely on heuristic or external schema linking modules, SQL2NL ensures schema consistency by construction, significantly reducing linking-related errors and removing a common confounder in NL2SQL evaluation. Moreover, the generated natural language queries maintain syntactic fluency, semantic equivalence to the source SQL, and structural fidelity to the underlying schema.
Our experiments reveal that despite preserving semantics, \textbf{even after controlling for schema linking errors, NL2SQL models are still highly sensitive to linguistic variations}, underscoring the brittleness of current approaches and the importance of evaluating beyond exact-match metrics. 
In addition, we provide additional experiments on top of the existing literature, and provide Pass@K metric for both NL2SQL and SQL2NL tasks.
With no surprise, with increasing K Pass@K metric improves for both tasks; providing a more robust metric for evaluating NL2SQL and SQL2NL by excluding random errors in LLM generation. With a high enough K, Pass@K metric reveals the maximum ability of an LLM, where we show that SQL2NL can outperform NL2SQL with big enough K.
Our approach provides a novel, scalable pathway for diagnosing and addressing failure modes in NL2SQL systems—paving the way for more resilient and generalizable benchmarks.
\section{Evaluating NL2SQL via SQL2NL}
\label{sec:method}

\subsection{Workflow Overview}

\begin{figure*}[t]
\vspace{-15pt}
\centering
\includegraphics[width=0.9\textwidth]{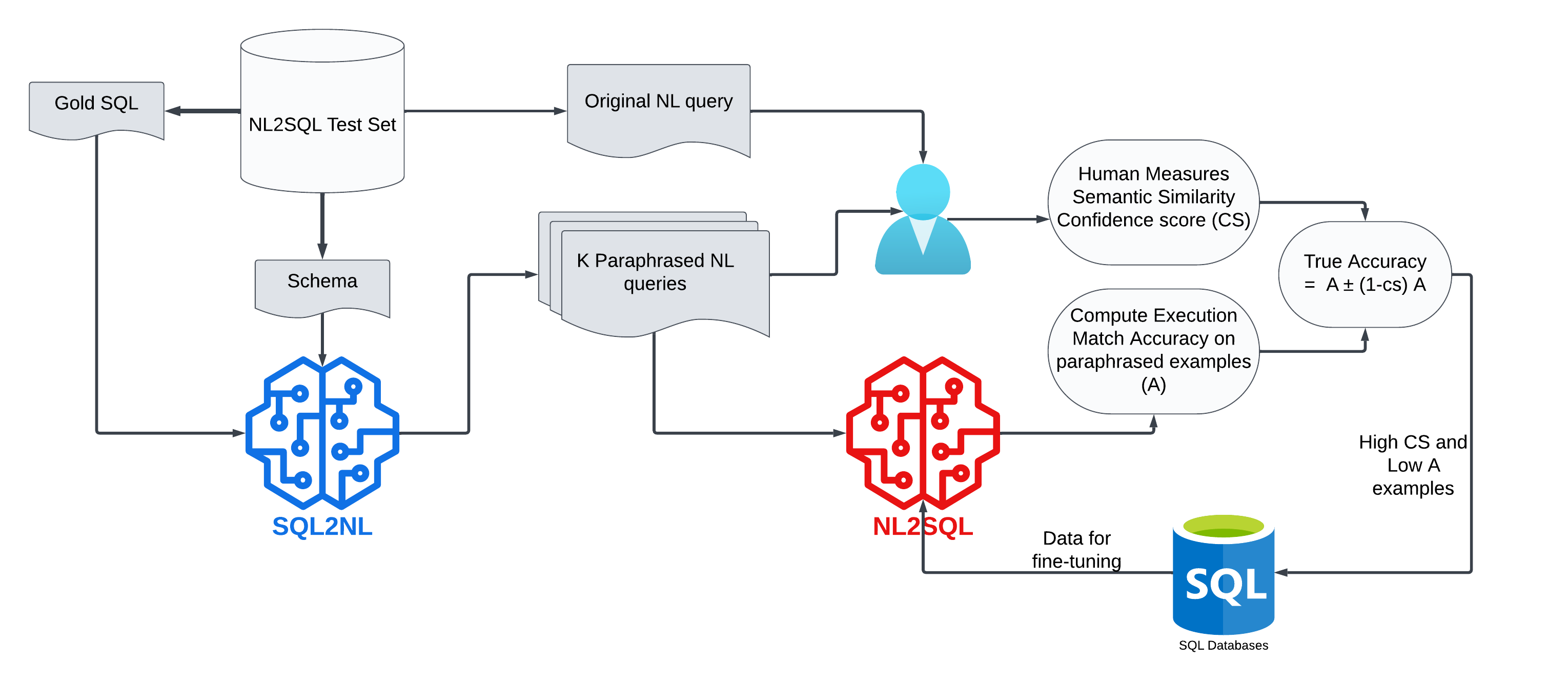}
\caption{We begin by extracting the gold SQL query and its associated schema from a benchmark NL2SQL test set. A SQL2NL model is then used to generate $k$ paraphrased natural language queries. Each paraphrase is semantically validated against the original query to ensure meaning preservation. The validated paraphrases are subsequently fed into the NL2SQL model to produce predicted SQL queries. Robustness is measured using execution-match accuracy across the paraphrases. While SQL2NL and NL2SQL can in principle be treated as separate modeling tasks, in this study we employ a single unified model for both directions.}
\label{fig:flowchart}
\vspace{-15pt}
\end{figure*}

Figure~\ref{fig:flowchart} shows the end-to-end pipeline for evaluating NL2SQL model robustness. We start by extracting gold SQL queries and schema from a test set. By grounding on the schema, a SQL2NL model then generates $k$ paraphrased NL queries that introduce linguistic variation while preserving semantic equivalence. \textbf{While SQL2NL and NL2SQL could be handled by separate models, here we use a single unified model for both.}

The NL2SQL model is then evaluated on paraphrased queries, measuring performance degradation against the original. Human evaluators assess semantic similarity, generating confidence scores ($CS$) to adjust accuracy for robust evaluation.

\subsection{Paraphrase Generation}

We generate paraphrased NL queries using a structured prompt template, designed to ensure linguistic diversity while preserving logical equivalence. The prompt combines schema definitions and SQL queries, followed by specific instructions to enforce diversity constraints. The full prompt template is provided in Appendix~\ref{app:prompt}.  

This approach is grounded in minimizing schema linking errors by leveraging SQL queries (\(G\)) and schema information (\(S\)) to generate paraphrased queries (\(Q\)) that exhibit stronger schema alignment than the original queries (\(Q_{orig}\)). Conditioning on both (\(G\) and \(S\) is expected to yield paraphrases that are semantically equivalent to the gold SQL while being explicitly grounded in the schema, thereby reducing schema mismatches. 

This assumption can also be expressed in terms of expected log-likelihood. 
Let $P_{\text{NL2SQL}}(Y \mid Q,S)$ denote the probability assigned by an 
NL2SQL model to a SQL query $Y$ given natural language query $Q$ and schema $S$, 
and let $Y_{\text{gold}}$ be the gold SQL. For the \textit{original dataset queries}, 
we treat them as an empirical distribution $P_{\text{data}}(Q_{\text{orig}})$, 
which often provides only a single natural language query per SQL. 
For \textit{paraphrased queries}, we instead define a distribution 
$P(Q \mid G,S)$ induced by a generator conditioned on the gold SQL $G$ and schema $S$.  

We claim that schema-grounded paraphrases improve the model’s objective:
\begin{equation}
\begin{aligned}
\mathbb{E}_{Q \sim P(Q \mid G,S)} 
   \big[ \log P_{\text{NL2SQL}}(Y_{\text{gold}} \mid Q,S) \big] \\
\quad>\;
\mathbb{E}_{Q \sim P_{\text{data}}(Q_{\text{orig}})} 
   \big[ \log P_{\text{NL2SQL}}(Y_{\text{gold}} \mid Q,S) \big]
\end{aligned}
\end{equation}

This inequality asserts that schema-conditioned paraphrases 
increase the expected log-likelihood of the correct SQL relative to the 
original queries. Intuitively, the gain depends on the nature of $Q_{\text{orig}}$: 
datasets with clear, unambiguous queries leave less room for improvement, 
while noisy or underspecified queries benefit substantially from paraphrasing. 
By embedding schema information directly into the paraphrasing process, 
the generated queries remain faithful to the logical intent of the gold SQL 
while introducing linguistic variation. These variations not only increase 
the likelihood of producing the correct SQL, but also stress-test NL2SQL models under diverse formulations, thereby exposing weaknesses in schema linking and improving robustness.

\subsection{Evaluation Metrics}

We evaluate performance degradation by comparing accuracy on paraphrased queries ($A_{para}$) against original queries ($A_{orig}$). Accuracy is measured using execution match (EM) accuracy, which evaluates whether the SQL query execution result matches the result of the gold SQL query when executed on the same database. EM accuracy is widely regarded as a more robust metric than string-based comparisons because it directly reflects whether the generated SQL produces the correct output, regardless of syntactic differences. This makes it particularly suitable for evaluating NL2SQL systems, where queries may vary in structure but yield identical results.  

To quantify performance degradation, we compute the accuracy drop as $\Delta_{acc} = A_{orig} - A_{para}$, where $A_{orig}$ and $A_{para}$ represent the execution match accuracies on the original and paraphrased queries, respectively. The paraphrased queries are expected to preserve semantic equivalence with the original NL queries. To account for this, we incorporate human evaluations of paraphrase quality. Annotators assess the semantic similarity between the original and paraphrased queries and provide confidence scores ($CS$) on a scale of 0 to 1, where higher scores indicate greater similarity. We use these confidence scores to adjust the measured accuracy, accounting for potential paraphrasing errors: $A_{true} = A_{para} \pm (1 - CS) \cdot A_{para}$


This adjustment ensures that errors arising from paraphrasing artifacts do not unfairly penalize model performance, providing a more reliable evaluation of robustness to linguistic variations. By combining execution match accuracy with human-validated paraphrase quality, this approach provides a rigorous framework for analyzing model sensitivity to variations in query phrasing, enabling reliable assessments in NL2SQL tasks.  

In this work we explore replacing the human in the loop with semantic similarity score from the embeddings and show the reliability of this approach for the datasets we study in this work.

\section{Nature of Paraphrased Queries}
\label{sec:nature_of_syn_queries}
\subsection{Semantic Similarity}
\begin{figure}[h]
    \centering
    \includegraphics[width=1\linewidth]{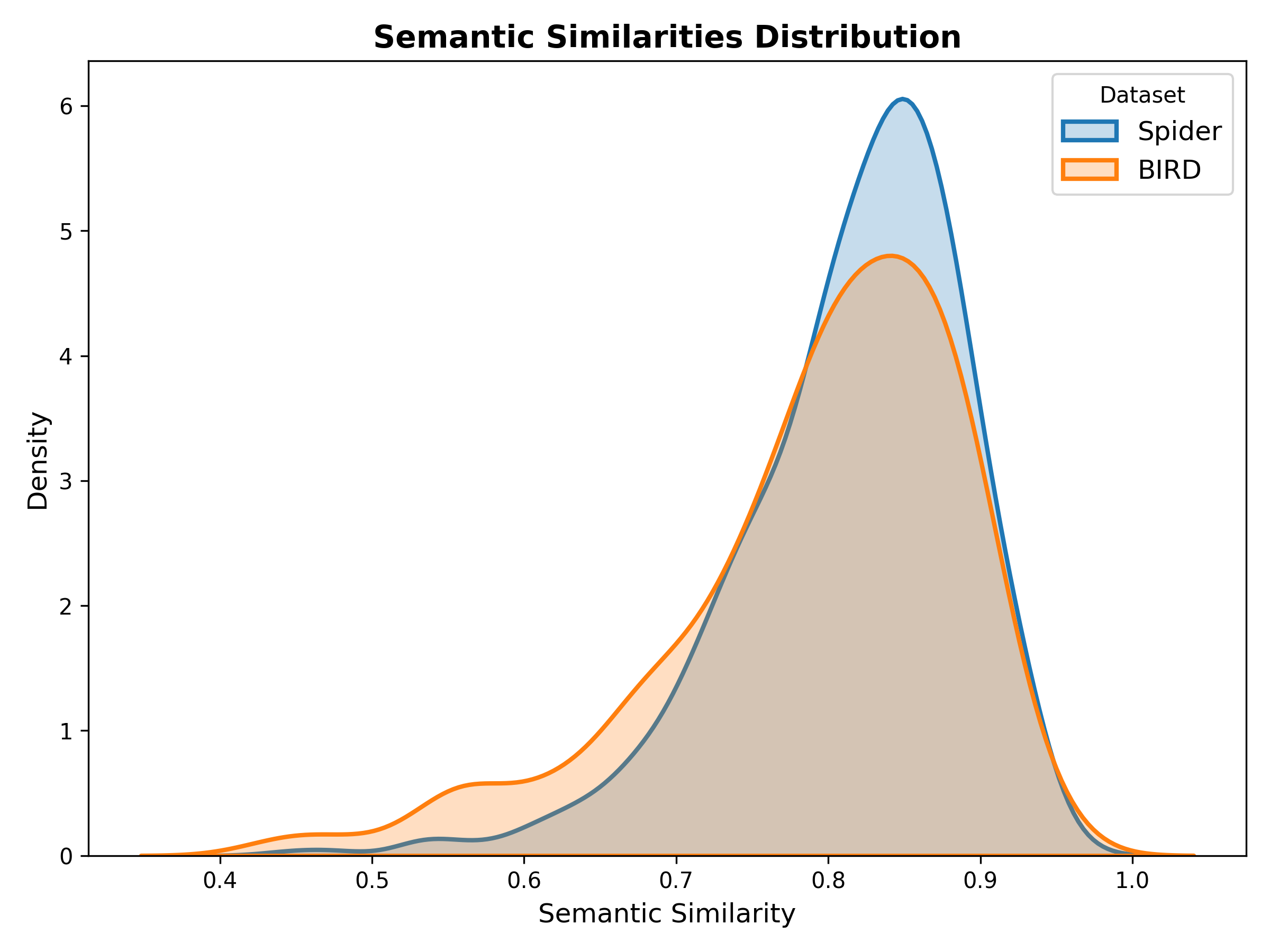}
    \caption{Kernel Density Estimation (KDE) plot of semantic similarities between original and paraphrased queries for the Spider and BIRD datasets using Sentence-
BERT embeddings. The Spider dataset exhibits a higher mean similarity (0.81) than BIRD dataset (0.79) and wider variance.}
\label{fig:semantic_similarity}
\vspace{-15pt}
\end{figure}

We analyze the semantic similarity between the original and paraphrased queries using Sentence-BERT embeddings, which measure similarity in a high-dimensional semantic space. The results presented in Figure \ref{fig:semantic_similarity} indicate a high degree of alignment, with a mean similarity score of 0.81 and a median score of 0.83 for the Spider dataset. The interquartile range spans from 0.77 to 0.86, suggesting that most paraphrases preserve the semantic intent of the original queries. Notably, the minimum similarity score is 0.44, indicating a subset of paraphrases that deviate more significantly from the original query. For the BIRD dataset, the mean similarity score is 0.8, with a wider variance. This broader distribution suggests that paraphrases in BIRD exhibit more linguistic diversity, potentially influenced by its more complex schema structures.

Table \ref{tab:semantic_examples} in Appendix~\ref{app:paraphrase_example} presents examples comparing the original queries with their paraphrased variants, along with semantic similarity scores. The examples illustrate varying levels of similarity, with scores ranging from 0.53 to 0.84. Through expert annotation, we observe that similarity scores above 0.6 correspond to paraphrases that accurately preserve the meaning and intent of the original query while introducing linguistic variation. In particular, queries with scores above 0.6 exhibit consistent semantic alignment with the original, even when rephrased using different sentence structures, word order, or synonyms.
Based on this threshold, \textbf{more than 98\% of the samples in our dataset are paraphrased reliably}, demonstrating the effectiveness of the paraphrasing process. The lower-scoring example (
0.53) highlights cases where subtle differences in emphasis or additional qualifiers may lead to minor deviations from the original query’s intent. However, even in such cases, the paraphrases remain interpretable and contextually relevant.
The results suggest that the generated paraphrases effectively test the robustness of NL2SQL models while maintaining semantic equivalence, enabling rigorous evaluations under input variations. 


\subsection{Grammatical Similarity Analysis}
\label{app:grammar}
We begin with evaluating the grammatical similarity between paraphrased and original natural language queries. We employed syntactic tree-based representations to measure structural alignment. Using the SpaCy library \citep{spacy2}, dependency parse trees were generated for each query, and their hierarchical structures were compared to compute a grammar similarity score. This metric quantifies the degree of syntactic overlap, capturing both grammatical consistency and structural variations.  

The grammatical similarity score (
$S_{grammar}(Q_1, Q_2)$
) was computed as the average of two components:

1. Tree Structure Similarity ($S_{tree}$): Measures the overlap in dependency parse trees, based on the number of subtrees. 
\begin{equation}
S_{tree}(Q_1, Q_2) = 1 - \frac{|T_1 - T_2|}{\max(T_1, T_2)}
\end{equation}
where $T_1$ and $T_2$ are the number of dependency subtrees in the queries.

2. POS Tag Similarity ($S_{pos}$): Computes overlap in part-of-speech (POS) tags.
\begin{equation}
S_{pos}(Q_1, Q_2) = \frac{|P_1 \cap P_2|}{\max(|P_1|, |P_2|)}
\end{equation}
where $P_1$ and $P_2$ represent the sets of POS tags in each query. The final grammar similarity score combines the two measures as:
\begin{equation}
S_{grammar} = \frac{S_{tree} + S_{pos}}{2}
\end{equation}

\begin{figure}[ht!]  
\centering  
\includegraphics[width=\columnwidth,height=5cm]{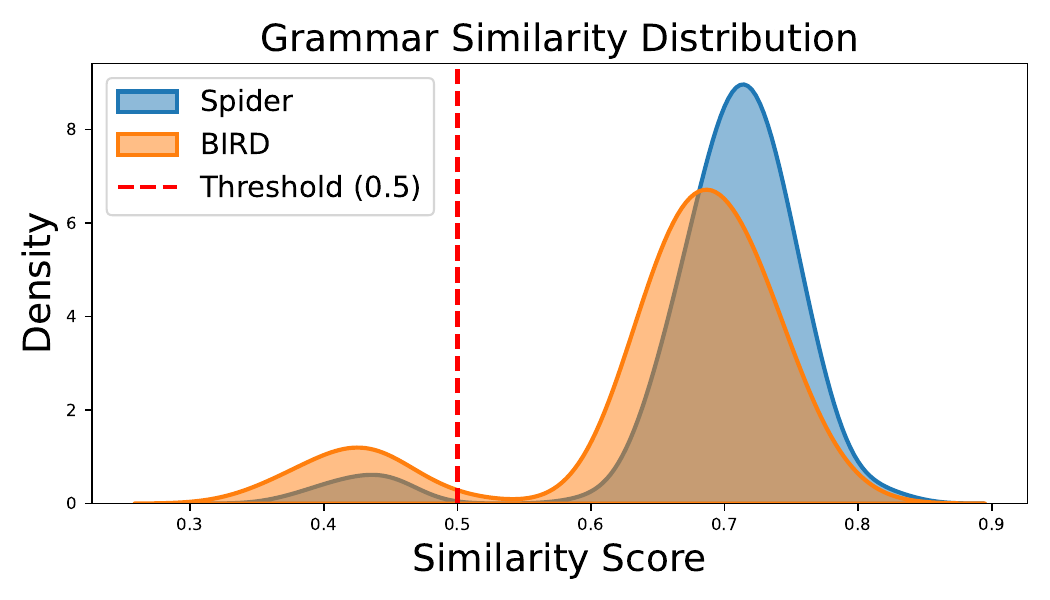}  
\caption{Grammatical similarity score distributions for the Spider and BIRD datasets. The bimodal patterns reflect cases of significant syntactic rephrasing (lower peak) and close syntactic alignment (higher peak).}  
\label{fig:grammar_similarity}  
\vspace{-15pt}
\end{figure}

We analyzed the grammatical similarity between the original queries and their paraphrased counterparts to investigate structural variations introduced during paraphrasing. Grammatical similarity was computed using dependency parse trees and POS tag alignments, capturing syntactic consistency while accounting for structural transformations. Figure~\ref{fig:grammar_similarity} presents the distribution of grammatical similarity scores for the Spider and BIRD datasets. The results reveal a bimodal distribution with two peaks: a higher similarity peak indicating paraphrases that closely match the syntax of the originals, and a lower similarity peak suggesting more substantial syntactic deviations.  

The mean grammatical similarity is 0.7 for Spider and 0.65 for BIRD, respectively. Both datasets exhibit a 95\% confidence interval (CI) spanning approximately [0.43, 0.76], indicating considerable variation in syntactic alignment. Several paraphrased queries exhibit low grammatical similarity scores due to structural differences from the original queries. Original queries often contain multi-clause formulations or explicit schema references, while paraphrased versions simplify structures while preserving semantic intent. For example, an original query:  
\textit{"How many countries does each continent have? List the continent id, continent name, and the number of countries."}  
is paraphrased as:  
\textit{"What is the distribution of countries across different continents?"} 


These transformations reduce syntactic complexity, alter POS tag patterns, and flatten dependency structures, resulting in lower grammar similarity scores despite semantic equivalence.

\subsection{Analysis of Syntactic and Lexical Features}
\label{app:syntactic}
We analyzed the syntactic and lexical properties of the original natural language queries and their
paraphrased counterparts to better understand the structural variations introduced during paraphrasing. Specifically, we examined sentence length, syntactic depth, and lexical diversity across the two sets of queries. Sentence length was measured by the number of tokens. Syntactic depth was computed as the maximum dependency distance between a word and its syntactic head:
\[
D_{syntactic}(w_i) = \max (|h(w_i) - i|)
\]
where \( h(w_i) \) denotes the index of the syntactic head of word \( w_i \) in the dependency parse tree, and \( i \) is the position of the word in the query. Lexical diversity was calculated as the ratio of unique words to the total number of words in a query, $LD = \frac{V}{N}$, where \( V \) represents the number of unique words and \( N \) represents the total number of words in the query. These metrics provide insights into the structural and lexical variations between the original and paraphrased queries, allowing us to assess whether paraphrasing preserves or alters query complexity and linguistic richness.

\begin{figure}[ht!]
\centering
\includegraphics[width=\columnwidth]{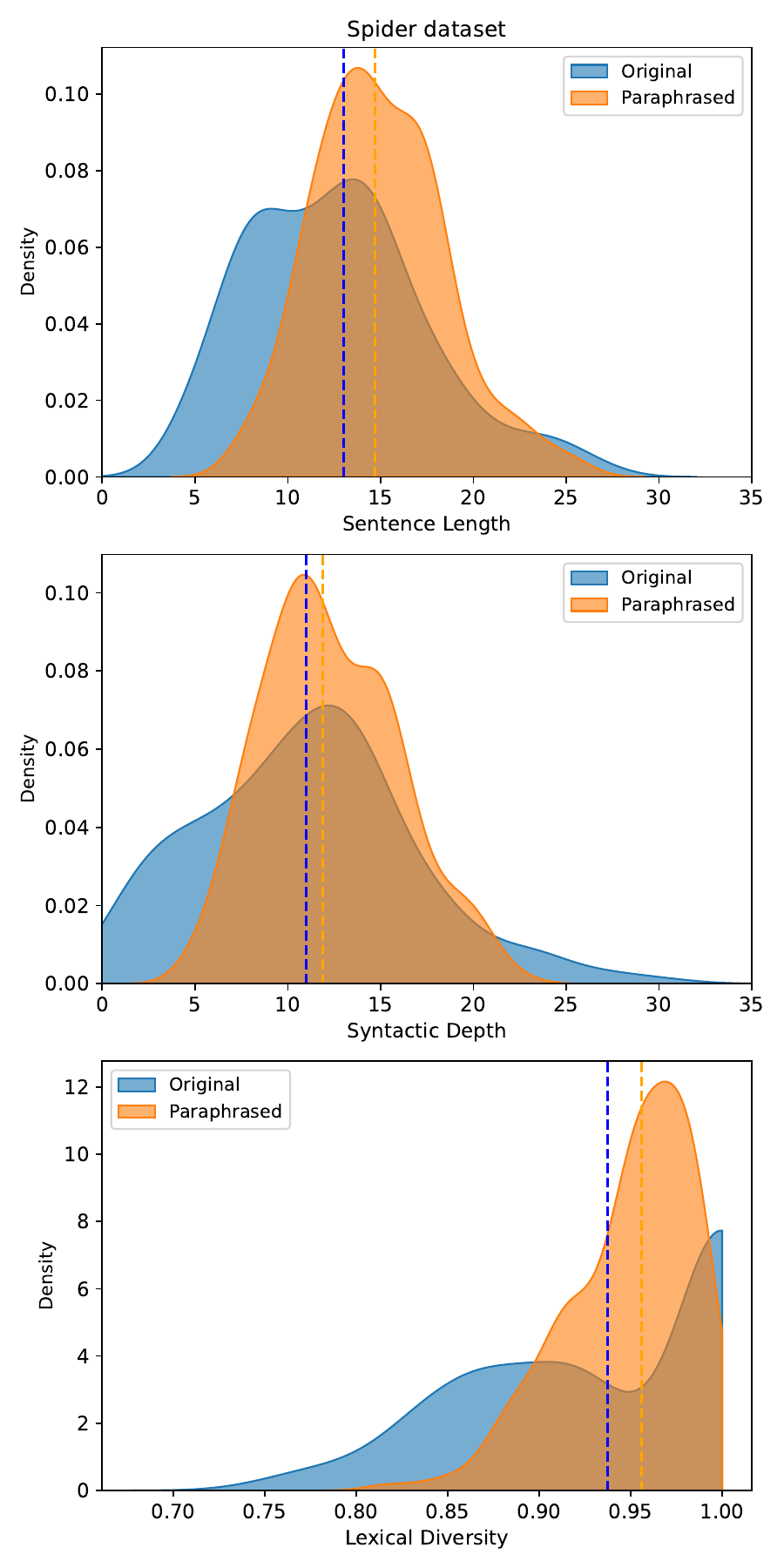}
\vspace{-15pt}
\caption{Distributions of sentence length, syntactic depth, and lexical diversity for original and paraphrased queries. Paraphrased queries are longer and exhibit slightly higher syntactic depth, while lexical diversity remains high and comparable to the original queries.}
\label{fig:syn_spider}
\vspace{-15pt}
\end{figure}

Figure~\ref{fig:syn_spider} visualizes the distributions of these features using kernel density estimation (KDE) plots for both the original and paraphrased queries of Spider datasets. Paraphrased queries are longer (15.02 vs. 12.4 tokens) and exhibit slightly greater syntactic depth (12.53 vs. 10.8), suggesting increased complexity. Lexical diversity remains high (0.94 for both), preserving vocabulary richness. However, paraphrased queries show lower variability in word choice, as indicated by a slightly reduced standard deviation in lexical diversity.
We present the result for BIRD dataset in Appendix~\ref{app:syntactic}.

The results for both Spider and BIRD datasets suggest that the paraphrased queries maintain grammatical and lexical richness while introducing structural complexity, making them suitable for testing the robustness of NL2SQL models against linguistic variations. 

\section{Experimental Results}
We evaluate LLM performance on paraphrased queries by randomly selecting 1000 examples from the Spider development set. For each gold SQL query \(Q_i\) (with its corresponding schema), we ask SQL2NL model to generate \(m = 10\) natural language queries, resulting in a total of 10,000 paraphrased examples, represented as pairs
\[
\{(N_1, Q_i), (N_2, Q_i), \ldots, (N_m, Q_i)\}.
\]

We test recent \textit{off-the-shelf} LLMs (including GPT-4o~\citep{achiam2023gpt} and LlaMa3-series of models~\citep{dubey2024llama}) on the paraphrased natural language queries. A detailed comparison of dataset domains, sizes, table characteristics, and query complexities is provided in Table \ref{tab:comparison_stats}.

\begin{table*}
\centering
\caption{Average schema and query complexities of NL2SQL datasets in this work.}
\renewcommand{\arraystretch}{1.1}
\resizebox{\textwidth}{!}{
\begin{tabular}{l|l|c|c|c|c|c|c|c}
\toprule
\textbf{Dataset} & \textbf{Domain} & \textbf{\#Queries} & \textbf{\#DB} & \textbf{\#Tables/DB} & \textbf{\#Cols/Table} & \textbf{\#Joins/Query} & \textbf{\#Agg/Query} & \textbf{Nest Depth/Query} \\
\midrule
Spider (Dev) & Misc.  & 1034 & 20 & 4.05 & 5.44 & 0.51  & 0.85 & 1.09 \\
Bird (Dev) & Misc. & 1534 & 11 & 6.82 & 10.6 & 0.92 & 0.66 & 1.09 \\
FIBEN & Financial & 300 & 1 & 41 & 3.71 & 4.1  & 1.41 & 1.60 \\
\bottomrule
\end{tabular}}
\label{tab:comparison_stats}
\end{table*}

\begin{table*}
\centering
\caption{Mean accuracy of off-the-shelf LLMs on original and paraphrased NL queries in the Spider dev set. Accuracy is in percentage. The "Degradation" column quantifies the absolute drop in accuracy due to paraphrasing. The same LLM is used for paraphrasing in each row.}
\vspace{-5pt}
\begin{adjustbox}{width=.85\textwidth}
\renewcommand{\arraystretch}{0.5} 
\begin{tabular}{l|c|c|c}
\toprule
\textbf{Model} & \textbf{Original Accuracy (\%)} & \textbf{Paraphrased Accuracy (\%)} & \textbf{Degradation (\%)} \\
\midrule
Llama3.1 405B   & 79.5  & 69.2  & \textbf{10.3 ↓}  \\
Llama3.3 70B    & 77.1  & 66.9  & \textbf{10.2 ↓}  \\
Llama3.1 8B     & 62.9  & 42.5  & \textbf{20.4 ↓}  \\
GPT4o-mini (8B) & 77.4  & 65.2  & \textbf{12.2 ↓}  \\
\bottomrule
\end{tabular}
\end{adjustbox}
\label{tab:llm_comparison}
\end{table*}

\begin{figure}[ht]
\centering
\includegraphics[width=.8\linewidth,height=6cm]{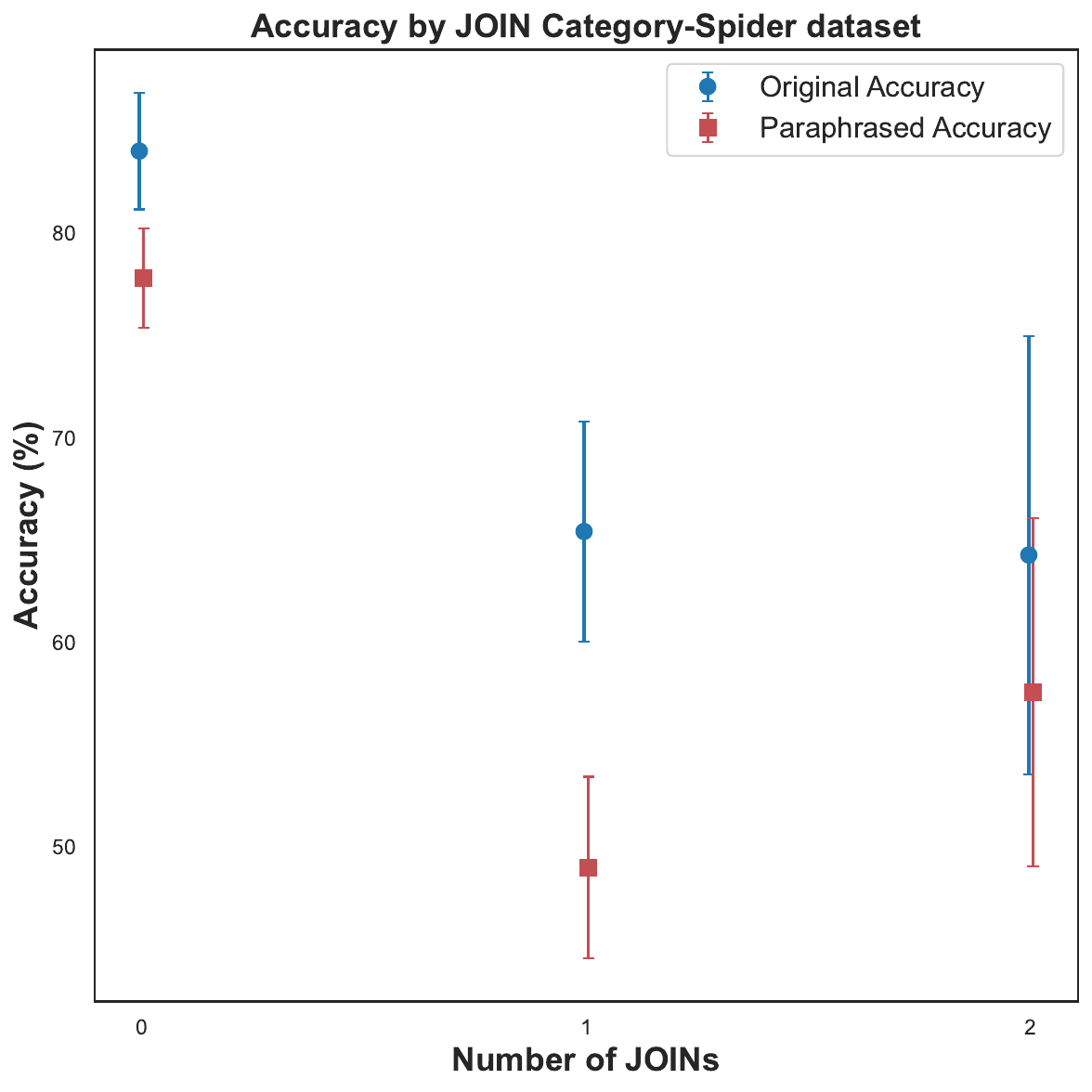}
\caption{Accuracy of LlaMa3.3-70B across different JOIN categories with bootstrap-based error bars for Spider dev set. Results are shown for paraphrased and original NL queries, base on execution match accuracy. Error bars represent 95\% confidence intervals, accounting for variability due to dataset size in each category.}
\vspace{-18pt}
\label{fig:compare_join_spider}
\end{figure}

\subsection{Comparing LLMs on Paraphrased and Original Queries}
\label{sec:different_models}
Our key experiments are designed to answer the following research question: are existing NL2SQL solution robust to paraphrases in the input NL queries. We compare the performance of a few off-the-shelf LLMs on the Spider dev set.
Given the results in Table ~\ref{tab:llm_comparison}, it is evident that the models exhibit varying levels of performance on paraphrased vs original queries. GPT4o-mini achieves the highest accuracy on original queries (77.4\%) but sees a moderate drop when tested on paraphrased queries (65.2\%), highlighting some sensitivity to input rephrasing. Similarly, Llama3.3 70b performs comparably to GPT4o-mini on original queries (77.1\%) and demonstrates slightly better robustness to paraphrasing, with a smaller drop to 66.9\%. In contrast, Llama3.1 8b not only shows lower overall performance but also experiences a much larger drop between original (62.9\%) and paraphrased queries (42.50\%), \textbf{with a decrease of over 20 percentage points.} This pronounced decline suggests that the smaller model struggles significantly with generalizing to paraphrased inputs, likely due to its limited capacity to handle the complex semantic mappings required for NL2SQL tasks.

These results highlight the critical trade-offs between model size, accuracy, and robustness to input variability. While larger models like Llama3.3 70b and GPT4o-mini exhibit better generalization, smaller models like Llama3.1 8b face significant challenges, particularly with paraphrased inputs. Addressing these gaps, especially for smaller models, could be key to improving performance and ensuring robustness in real-world applications where paraphrased inputs are common.

We estimate that the total computational budget required to obtain these results is approximately 1,000 GPU-hours on A100 GPUs.

\subsection{Performance by JOIN Count}

As a proxy for real-world complexity measure, 
we analyze performance based on the number of \texttt{JOIN} operations in the Spider dev set. Accuracy declines as \texttt{JOIN} complexity increases. For 0 \texttt{JOINs} , accuracy drops from 84.04\% (original) to 77.83\% (paraphrased), a 6.21\% decrease, indicating even simple paraphrases affect generalization. With 1 JOIN, accuracy falls from 65.44\% to 49\% (16.44\% drop), highlighting schema linking challenges. For 2 \texttt{JOINs} , accuracy declines from 64.29\% to 57.57\% (6.72\% drop), suggesting overestimated accuracy for simpler queries. The result is shown in Figure~\ref{fig:compare_join_spider}.

\begin{figure}[t]
\centering
\includegraphics[width=\columnwidth]{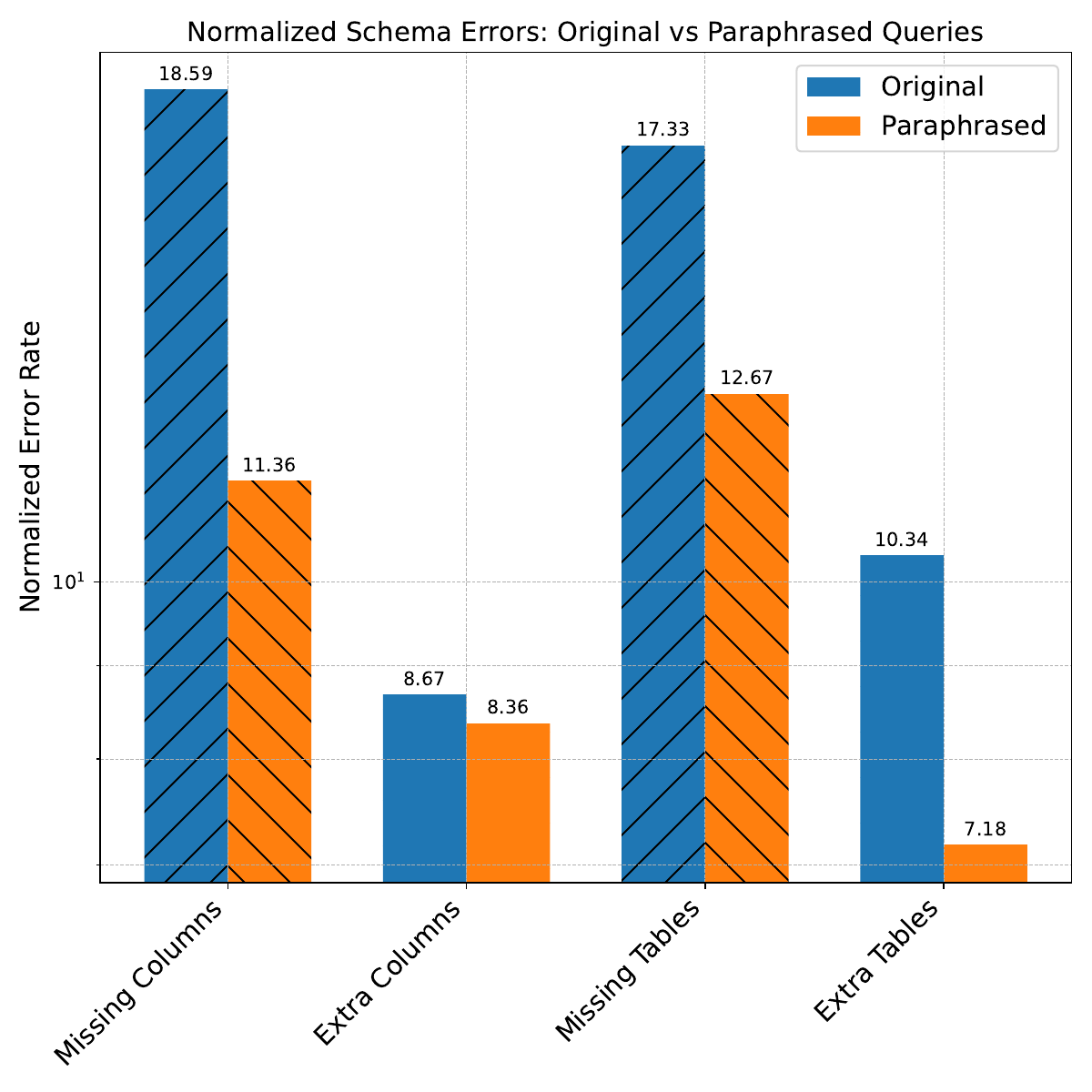}
\caption{Normalized schema error rates for \textit{original} and \textit{paraphrased queries}. Errors in \textit{False cases} are scaled relative to \textit{True case baselines} to ensure fair comparison. Paraphrased queries exhibit lower normalized error rates across all categories, suggesting improved schema alignment.}
\label{fig:normalized_errors}
\vspace{-22pt}
\end{figure}

We extend the analysis of accuracy trends concerning the number of \texttt{JOINs}  by incorporating the FIBEN dataset \citep{FIBEN_dataset}, which features a significantly higher average number of \texttt{JOINs}  per query compared to Spider and BIRD. Detailed numerical results for both BIRD and FIBEN are discussed in~\ref{app:accuracy_joins}, Tables 4, 5, and 6. We observe that Spider shows a sharper accuracy decline with complexity, while BIRD is more resilient. For FIBEN we observe no clear trend.

\subsection{Clause-based performance analysis}
We analyze the impact of paraphrased NL queries on model performance as a function of SQL clause presence in Appendix~\ref{app:accuracy_clause}. Our findings indicate that the model experiences greater degradation on Spider, particularly for queries involving \texttt{ORDER BY}, \texttt{GROUP BY}, and nested queries, while BIRD demonstrates more stable performance with minimal degradation across most clauses. This suggests that the model struggles more with paraphrasing in datasets where SQL queries exhibit higher structural complexity and a stronger dependence on precise linguistic patterns.

\subsection{Normalized Schema Error Analysis}

We evaluated schema alignment errors in both original and paraphrased queries by analyzing four error types: \textit{missing columns}, \textit{extra columns}, \textit{missing tables}, and \textit{extra tables}. Using the SQLGlot library, we parsed SQL queries to extract schema elements while accounting for alias resolution, projection order differences, and implicit projections. Errors were computed by comparing predicted and gold schema elements. To ensure a fair comparison, error rates in False cases were normalized relative to true case baselines. We formulate normalization as follows:
\begin{equation}
NER = \frac{E_{false}}{E_{true} + \epsilon}
\end{equation}
where $E_{false}$ and $E_{true}$ represent the error rates in False and True cases, respectively, and $\epsilon$ is a small constant added to avoid division by zero. This approach controls for differences in baseline complexity, enabling meaningful comparisons.

The results, visualized in Figure~\ref{fig:normalized_errors}, reveal that \textbf{due to controlling for schema alignment, normalized error rates are consistently lower for paraphrased queries compared to original queries across all error categories.} On average, normalized errors for paraphrased queries (9.9) are significantly lower than those for original queries (13.7). This trend is observed in both column-related and table-related errors.

These findings suggest that schema alignment errors, particularly those arising from implicit \texttt{JOINs}  and nested structures, are reduced when queries are paraphrased. One possible explanation is that paraphrased queries, generated from SQL and schema information, enforce more explicit references to schema elements, thereby minimizing schema linking errors. Additionally, the reduced complexity observed in paraphrased queries likely improves alignment with the schema, further mitigating errors. While paraphrasing reduces schema alignment errors overall, some remaining errors highlight the need for improved schema-linking mechanisms to handle intricate query patterns effectively.

\section{Pass@K Performance}
As shown in previous section, and mentioned in \cite{bhaskar-etal-2023-benchmarking}, there is a common belief that parameterized queries ends in dropped performance. We further expanded the experiments for the GPT4o-8B model on Spider dataset by running 10 replica of each query to obtain Pass@K on both the SQL2NL and NL2SQL tasks. 
As shown in Table~\ref{tab:pass_at_k_result}, with no surprise, increasing K ends in higher performance for both tasks, and contrary to the common belief that parameterized queries always ends in dropped performance, SQL2NL outperforms NL2SQL for $K \in \{5, 10\}$. Also, as shown for both tasks, the performance gain reaches to the boundaries of LLM capability, such that there is only 1\% improvement from Pass@5 to Pass@10. Details of our experiments is discussed in Appendix \ref{app:pass_at_k}.

\begin{table}[]
\centering
\caption{The Pass@K value for Spider dev dataset}
\begin{adjustbox}{width=.35\textwidth}
\begin{tabular}{l|llll}
K      & 1      & 2      & 5      & 10     \\ \hline
nl2sql & 76.0\% & 78.6\% & 80.6\% & 81.3\% \\
sql2nl & 67.1\% & 76.5\% & 83.3\% & 84.6\%
\end{tabular}
\label{tab:pass_at_k_result}
\end{adjustbox}
\vspace{-15pt}
\end{table}

\section{Beyond Evaluation: Leveraging SQL2NL for NL2SQL Training} \label{sec:conclusion}
In this study, we introduced a controlled framework for evaluating NL2SQL model robustness via semantically equivalent paraphrased queries. A key contribution is the use of SQL2NL to generate natural language queries that preserve the intent of the original SQL \textbf{while explicitly minimizing schema linking errors}. This isolation of schema alignment effects enables more precise error analysis and a clearer understanding of model failure modes.
Our results indicate that \textbf{even after controlling for schema linking errors, linguistic variations can lead to significant performance degradation—e.g., a 20\% accuracy drop for LLaMa3.1-8B on the Spider dev set}—underscoring the sensitivity of current models to paraphrasing and the need for more resilient solutions.

Beyond evaluation, this framework can support targeted training of NL2SQL models—building on approaches like CodeS \citep{CodeS}—via contrastive or adversarial fine-tuning. By focusing on failure cases where performance drops despite schema-consistent semantics, adopting this approach can be used to construct high quality preference datasets for further training of NL2SQL models and driving robustness in real-world deployments.

\section*{Limitations}
This work does not yet address more complex natural language queries, nor does it explore ambiguous or unanswerable queries. Additionally, the performance of models on multi-turn dialogues remains unexplored. These more challenging scenarios are closer to real-world applications and represent a crucial next step for improving our approach. Incorporating such cases into the evaluation pipeline will help drive the development of more robust NL2SQL systems, better equipped to handle the variety and complexity of real-world queries.

Moreover, despite focusing on the NL2SQL task under the assumption that the correct schema is provided, we acknowledge that real-world scenarios often involve schema retrieval as a critical pre-processing step. The ability to identify and retrieve the relevant schema remains an unresolved challenge, as noted by \citet{2024arXiv240902038B,2024arXiv240409889B}. Extending this evaluation framework to incorporate schema retrieval will provide a more comprehensive assessment of model robustness, addressing the interplay between schema matching and query generation in practical applications.

\section*{Ethics Statement}
This work does not pose any ethical issues. All the
data used in this paper are publicly available and
are used under following licenses: Creative Commons BY-SA 4.0 License, and Apache-2.0 license. Additionally, we utilized AI tools for proofreading and to identify potential grammar mistakes, ensuring clarity and coherence in the manuscript.

\bibliography{iclr2024_conference}
\appendix

\section{Prompt Template}
\label{app:prompt}
\begin{lstlisting}
Given the following database schema and an SQL query, generate {num_queries} distinct natural language questions that describe the purpose and output of the SQL query.

{schema_definitions}

SQL Query:
{sql_query}

Instructions:
1. Generate {num_queries} natural language questions that reflect the intent of the SQL query.
2. Each question should vary in phrasing, structure, and wording, but all questions must remain logically equivalent.
3. Do not include explanations, task descriptions, or any additional comments in the output.

Output Format:
1. <First question>
2. <Second question>
...
{num_queries}. <Nth question>.
\end{lstlisting}

\section{paraphrased query examples}
\label{app:paraphrase_example}
Table 3 shows examples of the original NL query from Spider dev set and their corresponding paraphrased queries generated through the SQL2NL framework.
\begin{table*}[t]
\centering
\caption{Examples comparing original queries with paraphrased variants and their semantic similarity scores. Semantic similarity was computed using Sentence-BERT (SBERT).}
\label{tab:semantic_examples}
\renewcommand{\arraystretch}{1} 
\setlength{\tabcolsep}{4pt} 
\resizebox{\linewidth}{!}{
\begin{tabular}{p{0.30\textwidth} p{0.5\textwidth} c}
\toprule
\textbf{Original Query} & \textbf{Paraphrased Variants (5 out of 10)} & \textbf{Similarity Score} \\ 
\midrule

\multirow{5}{=}{Find the name of the makers that produced some cars in the year of 1970?} 
& What car manufacturers produced models in 1970? & \multirow{5}{*}{0.84} \\
& Which car makers have models associated with vehicles from the year 1970? & \\
& What are the names of car makers that produced cars in 1970? & \\
& Can you list the car manufacturers whose models were released in 1970? & \\
& Which makers are associated with car models from 1970, according to the database records? & \\
\midrule

\multirow{5}{=}{What are the locations and names of all stations with capacity between 5000 and 10000?} 
& What are the names and locations of stadiums with capacities ranging from 5000 to 10000? & \multirow{5}{*}{0.68} \\
& Which stadiums have capacities between 5000 and 10000, and what are their names and locations? & \\
& How can I find the names and locations of stadiums that can accommodate between 5000 and 10000 people? & \\
& What stadiums, along with their locations, have a capacity of at least 5000 but no more than 10000? & \\
& Can you provide a list of stadium locations and names where the seating capacity is between 5000 and 10000? & \\
\midrule

\multirow{5}{=}{Return the grade that has the greatest number of high schoolers.} 
& What is the most common grade level among all high school students? & \multirow{5}{*}{0.60} \\
& Which grade has the largest number of students enrolled in it? & \\
& What grade level appears most frequently in the high school student database? & \\
& Can you identify the grade with the highest enrollment count? & \\
& What is the grade that the majority of high school students are in? & \\
\midrule

\multirow{5}{=}{What is the maker of the car produced in the earliest year and what year was it?} 
& What are the car makes that were produced in the earliest year available in the database? & \multirow{5}{*}{0.53} \\
& Which car manufacturers had models released in the minimum year listed in the cars data table? & \\
& What makes of cars were first introduced in the earliest year recorded in the database? & \\
& Can you list the car makes that correspond to the oldest year of production found in the cars data table? & \\
& What are the names of the car manufacturers that produced vehicles in the earliest year on record? & \\
\bottomrule
\end{tabular}}
\end{table*}

\section{Analysis of Syntactic and Lexical Features}
\label{app:syntactic}

\begin{figure}[ht!]
\centering
\includegraphics[width=\linewidth]{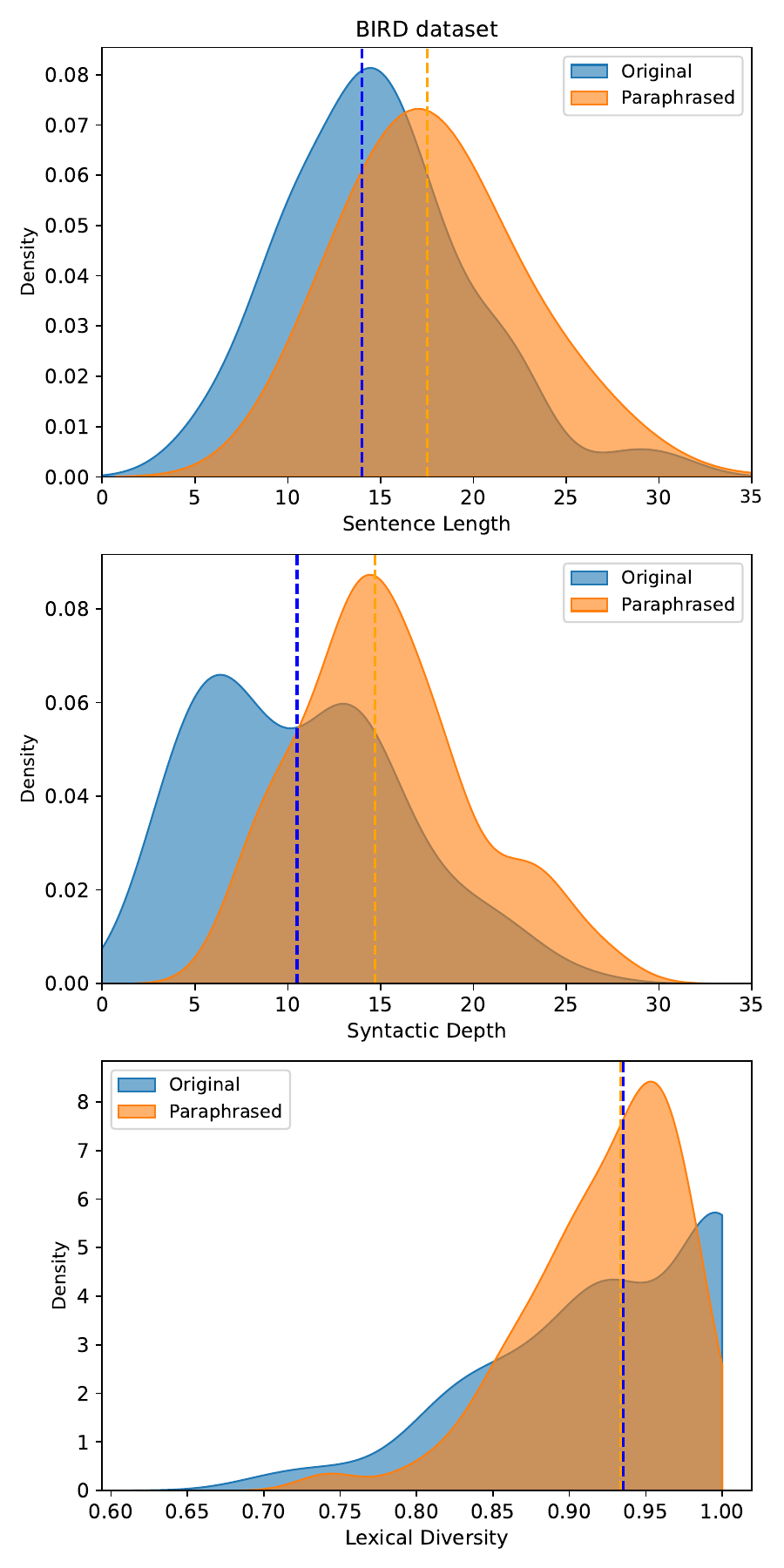}
\caption{Distributions of sentence length, syntactic depth, and lexical diversity for original and paraphrased queries. Paraphrased queries are longer and exhibit slightly higher syntactic depth, while lexical diversity remains high and comparable to the original queries.}
\label{fig:syntactic_bird}
\end{figure}

For the BIRD dataset, paraphrased queries are longer (18.53 vs. 14.54 tokens) and exhibit greater syntactic depth (15.68 vs. 10.70), indicating increased complexity. Lexical diversity is slightly lower in paraphrased queries (0.92 vs. 0.94), though the difference is minimal, preserving vocabulary richness. See Figure~\ref{fig:syntactic_bird}.

\section{Accuracy as a function of Joins for BIRD dev set}
\label{app:accuracy_joins}

For the BIRD dataset, accuracy varies with query complexity. With 0 \texttt{JOINs} , the model performs similarly on original (59.51\%) and paraphrased (59.68\%) queries, showing robustness. With 1 JOIN, accuracy declines from 56.47\% to 52.13\% (4.34\% drop), suggesting schema linking errors. For 2 \texttt{JOINs} , paraphrased queries outperform original ones (45.23\% vs. 43.4\%, +1.8\%), indicating paraphrasing may aid schema alignment in complex queries, highlighting differences between BIRD and Spider datasets.

Our results indicate that for FIBEN the performance is similiar and within the errorbar for both paraphrased and original NL queries regardless of the number of \texttt{JOINs}. This shows that the impact of this parameter is highly sensitive to the database under study.

\begin{table*}[]
\centering
\caption{Accuracy by JOIN Count for Spider Dataset}
\renewcommand{\arraystretch}{1.2}
\setlength{\tabcolsep}{10pt}
\begin{tabular}{lccc}
\toprule
JOIN Count & Examples & Original (\%) & Paraphrased (\%) \\
\midrule
0 & 612 & 84.31 ± 2.80 & 77.76 ± 2.33 \\
1 & 298 & 65.44 ± 5.54 & 49.00 ± 4.36 \\
2 & 70  & 64.29 ± 10.23 & 57.57 ± 8.80 \\
3 & 10  & 80.00 ± 19.26 & 12.00 ± 13.93 \\
4 & 6   & 66.67 ± 33.43 & 33.33 ± 23.96 \\
\bottomrule
\end{tabular}
\end{table*}
\begin{table*}[]
\centering
\caption{Accuracy by JOIN Count for BIRD and FIBEN Datasets}
\renewcommand{\arraystretch}{1.2}
\setlength{\tabcolsep}{10pt}
\begin{tabular}{lccc}
\toprule
JOIN Count & Examples & Original (\%) & Paraphrased (\%) \\
\midrule
0 & 247 & 59.51 ± 6.20  & 59.68 ± 4.49  \\
1 & 571 & 56.57 ± 4.19  & 52.05 ± 3.13  \\
2 & 122 & 43.44 ± 8.97  & 45.23 ± 6.97  \\
3 & 14  & 35.71 ± 28.09 & 54.29 ± 22.94 \\
\bottomrule
\end{tabular}
\end{table*}
\begin{table*}[]
\centering
\caption{Accuracy by JOIN Count for FIBEN Dataset}
\renewcommand{\arraystretch}{1.2}
\setlength{\tabcolsep}{10pt}
\begin{tabular}{lccc}
\toprule
JOIN Count & Examples & Original (\%) & Paraphrased (\%) \\
\midrule
0 & 0  & 0.00 ± 0.00  & 0.00 ± 0.00  \\
1 & 68 & 48.53 ± 12.04 & 62.46 ± 8.55 \\
2 & 10 & 30.00 ± 29.69 & 38.00 ± 20.71 \\
3 & 58 & 56.90 ± 11.56 & 52.00 ± 8.93  \\
4 & 41 & 36.59 ± 15.02 & 28.38 ± 10.90 \\
5 & 4  & 50.00 ± 49.10 & 52.50 ± 37.27 \\
6 & 30 & 46.67 ± 16.70 & 6.16 ± 4.40   \\
7 & 9  & 22.22 ± 22.82 & 18.55 ± 11.07 \\
\bottomrule
\end{tabular}
\end{table*}

\begin{figure}[ht]
\centering
\includegraphics[width=\linewidth, keepaspectratio]{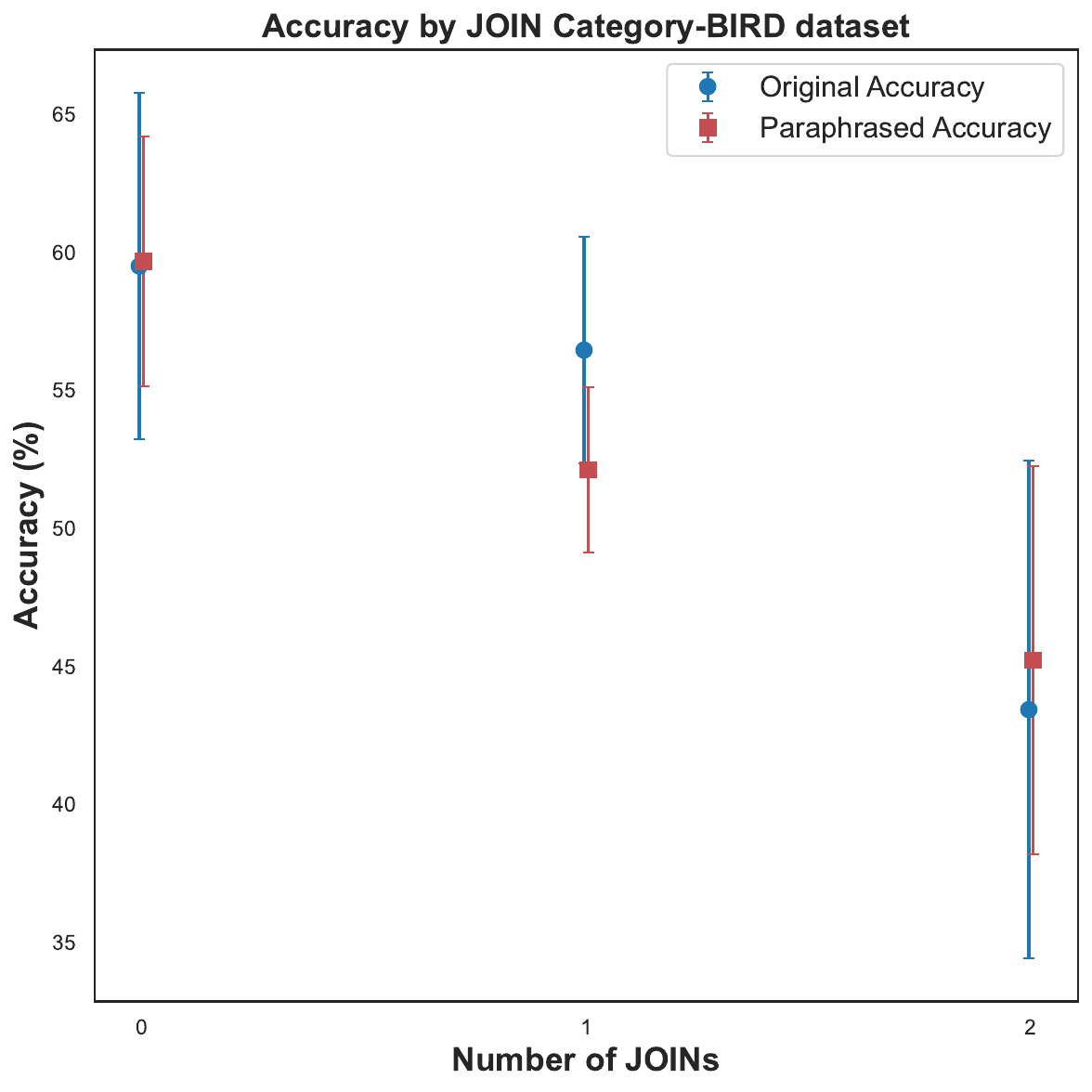}
\vspace{0.5em} 
\includegraphics[width=\linewidth, keepaspectratio]{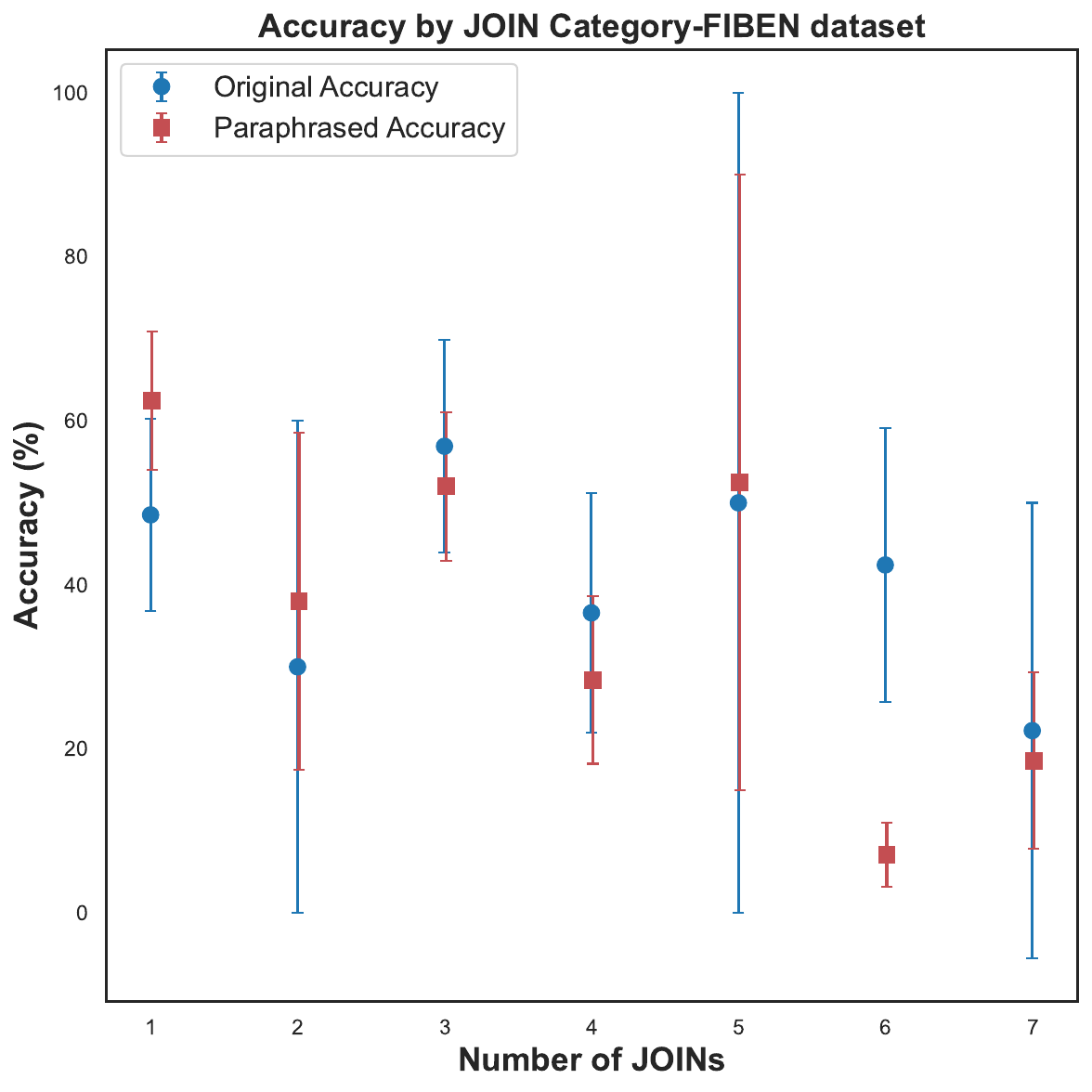}
\caption{Accuracy of LLaMa3.3-70B across different JOIN categories with bootstrap-based error bars for BIRD (top) and FIBEN (bottom). Results are shown for paraphrased and original NL queries, based on execution match accuracy. Error bars represent 95\% confidence intervals, accounting for variability due to dataset size in each category.}
\label{fig:compare}
\end{figure}

\section{Accuracy as a function of Clause presence}
\label{app:accuracy_clause}

Table~\ref{table:clause} shows the accuracies for queries with different SQL clauses. 
LlaMa-3.3 70B exhibits varying levels of degradation when handling paraphrased queries, with a stronger impact observed in cases where specific SQL clauses are present. For the \texttt{GROUP BY} clause, accuracy drops by 14.64\% (8.73\%) when the clause is present (absent), indicating that the model struggles more with paraphrased queries requiring explicit grouping. Similarly, the presence of the \texttt{ORDER BY} clause results in the most severe degradation, with accuracy decreasing by 27.93\% (5.24\%) in its presence (absence), highlighting a pronounced sensitivity to ordering operations. In the case of \texttt{HAVING}, the model unexpectedly shows a smaller degradation of 4.80\% (10.75\%) when the clause is present (absent), suggesting that it handles paraphrased conditions within the \texttt{HAVING} clause more robustly than conditions appearing in other parts of the query. For nested queries, the degradation is 12.69\% (9.87\%) in their presence (absence), further demonstrating the model's increased difficulty in handling paraphrased queries with complex query structures. Overall, the results indicate that degradation is more severe in the presence of the \texttt{ORDER BY}, \texttt{GROUP BY}, and nested queries, suggesting that the model is particularly vulnerable to paraphrasing in SQL queries that require explicit ordering, grouping, and structural complexity.

The model exhibits less severe degradation when handling paraphrased queries in the BIRD dataset compared to Spider. For the \texttt{GROUP BY} clause, accuracy drops by 6.93\% (1.81\%) when the clause is present (absent), indicating that explicit grouping impacts performance more under paraphrasing. The \texttt{ORDER BY} clause presents an unusual trend, where performance actually improves slightly in its presence, with a change of +1.75\% (3.14\%). The impact of paraphrasing on the \texttt{HAVING} clause is minimal, with a drop of just 1.00\% (2.15\%) when present (absent), suggesting robustness to paraphrased filtering conditions. For nested queries, degradation is 3.87\% (1.99\%) in their presence (absence), indicating that while paraphrasing affects complex query structures, the impact remains lower than that observed in Spider. Overall, the degradation is most pronounced for the \texttt{GROUP BY} and \texttt{Nested Queries} clauses, though the overall performance drop in BIRD remains more moderate than in Spider.

These results highlight the importance of evaluating clause-specific performance to identify areas where query translation models may require further improvements.

\begin{table*}[ht]
\centering
\caption{Accuracy by SQL Clauses for FIBEN, Spider, and BIRD Datasets (Original vs Paraphrased). The error bars are from bootstrap resampling.}
\makebox[\textwidth]{
\begin{adjustbox}{width=.8\textwidth}
\begin{tabular}{llcccccc}
\toprule
\multirow{2}{*}{Clause} & \multirow{2}{*}{Dataset} & \multirow{2}{*}{\#Examples} & \multicolumn{2}{c}{Original} & \multicolumn{2}{c}{Paraphrased} \\
\cmidrule(lr){4-5} \cmidrule(lr){6-7}
 & & & Acc (\%) & ± Err & Acc (\%) & ± Err \\
\midrule

\multirow{3}{*}{GROUP BY (Without)} 
  & FIBEN  & 149 & 52.35 & 8.01 & 49.39 & 6.05 \\
  & Spider & 731 & 82.49 & 2.59 & 73.76 & 2.34 \\
  & BIRD   & 895 & 56.87 & 3.06 & 55.06 & 2.56 \\
\midrule

\multirow{3}{*}{GROUP BY (With)} 
  & FIBEN  & 97  & 25.77 & 8.43 & 21.07 & 6.31 \\
  & Spider & 265 & 62.26 & 6.04 & 47.62 & 4.78 \\
  & BIRD   & 61  & 32.79 & 11.31 & 25.86 & 9.12 \\
\midrule

\multirow{3}{*}{ORDER BY (Without)} 
  & FIBEN  & 223 & 43.50 & 6.51 & 36.13 & 5.15 \\
  & Spider & 774 & 77.78 & 2.90 & 72.54 & 2.47 \\
  & BIRD   & 762 & 60.37 & 3.53 & 57.23 & 2.78 \\
\midrule

\multirow{3}{*}{ORDER BY (With)} 
  & FIBEN  & 23  & 26.09 & 17.31 & 59.57 & 11.78 \\
  & Spider & 222 & 74.77 & 6.24 & 46.84 & 4.97 \\
  & BIRD   & 194 & 35.57 & 7.21 & 37.32 & 4.71 \\
\midrule

\multirow{3}{*}{HAVING (Without)} 
  & FIBEN  & 182 & 45.05 & 7.03 & 46.37 & 5.54 \\
  & Spider & 920 & 78.91 & 2.72 & 68.16 & 2.39 \\
  & BIRD   & 946 & 55.60 & 3.10 & 53.45 & 2.61 \\
\midrule

\multirow{3}{*}{HAVING (With)} 
  & FIBEN  & 64  & 32.81 & 12.39 & 15.28 & 6.39 \\
  & Spider & 76  & 55.26 & 12.56 & 50.46 & 7.91 \\
  & BIRD   & 10  & 30.00 & 29.69 & 29.00 & 21.71 \\
\midrule

\multirow{3}{*}{Nested (Without)} 
  & FIBEN  & 129 & 51.94 & 8.62 & 57.35 & 6.19 \\
  & Spider & 844 & 78.79 & 2.86 & 68.92 & 2.57 \\
  & BIRD   & 890 & 56.52 & 3.09 & 54.53 & 2.42 \\
\midrule

\multirow{3}{*}{Nested (With)} 
  & FIBEN  & 117 & 30.77 & 7.92 & 17.30 & 5.21 \\
  & Spider & 152 & 67.76 & 7.13 & 55.07 & 5.62 \\
  & BIRD   & 66  & 39.39 & 12.01 & 35.52 & 9.23 \\

\bottomrule
\end{tabular}
\end{adjustbox}
\label{table:clause}
}
\captionsetup{width=.8\textwidth, justification=centering} 

\end{table*}

\section{Details of Pass@K Experiments}
\label{app:pass_at_k}
To get the Pass@K metric for NL2SQL, we used temperature of 0.5, and generated 10 different SQL queries with calling the same prompt. Then, either of the SQL commands are executed and the outcome is compared to the output of the gold SQL command. 

For SQL2NL, we passed the schema and the gold SQL query to a prompt, asking the LLM to generate 10 distinct text-queries. Then, each of the generated text-queries align with the schema are passed to the LLM to get the corresponding SQL queries. Finally, the SQL queries are evaluated with the same approach mentioned above.

We use the following formula to compute Pass@K metric:

$$E\left[ 1 - \frac{\binom{n - c}{k}}{\binom{n}{k}} \right].$$
where $n$ is the total, and $c$ is the total number of success.
\end{document}